\documentclass[journal]{IEEEtran}

\usepackage{hyperref}
\usepackage{graphicx}
\usepackage{subfigure}
\usepackage[flushleft]{threeparttable}
\usepackage{booktabs} % To thicken table lines
\usepackage{amsfonts}
\usepackage{multirow}
\usepackage{tabularx}
\usepackage{xcolor}
\usepackage{footnote}
\usepackage{caption}
\usepackage{textcomp, gensymb}
\usepackage{amsmath}

\newcolumntype{M}{>{$}c<{$}}
\newcolumntype{Z}{>{\centering\arraybackslash}X}
\newcolumntype{C}{>{\centering\arraybackslash}p}
\newcolumntype{Y}{>{\centering\arraybackslash}X}
\newcolumntype{L}{>{\raggedright\arraybackslash}p}

\newcommand{\specialcell}[2][c]{%
  \begin{tabular}[#1]{@{}c@{}}#2\end{tabular}}

\ifCLASSINFOpdf
\else
\fi

\begin{document}
\title{SpoofGAN: Synthetic Fingerprint Spoof Images}
% \title{SpoofGAN: Realistic Fake Fingerprint Spoof Images}

\author{Steven~A.~Grosz~and~Anil~K.~Jain,~\IEEEmembership{Life~Fellow,~IEEE}% <-this % stops a space
\thanks{S.A. Grosz and A.K. Jain are with the Department of Computer Science and Engineering, Michigan State University, East Lansing, MI, 48824 USA (e-mail: groszste@cse.msu.edu, jain@cse.msu.edu).}% <-this % stops a space
%\thanks{Manuscript received April 19, 2005; revised August 26, 2015.}
}

% The paper headers
\markboth{Journal of \LaTeX\ Class Files,~Vol.~14, No.~8, August~2015}%
% \markboth{Final Report for the NIST project ``Towards Synthesizing One Billion Fingerprints", 60NANB20D188, 10/1/20 to 9/30/21}
{Grosz \MakeLowercase{\textit{et al.}}: SpoofGAN: Synthetic Fingerprint Spoof Images}

% make the title area
\maketitle

% As a general rule, do not put math, special symbols or citations
% in the abstract or keywords.
\begin{abstract}
A major limitation to advances in fingerprint spoof detection is the lack of publicly available, large-scale fingerprint spoof datasets, a problem which has been compounded by increased concerns surrounding privacy and security of biometric data. Furthermore, most state-of-the-art spoof detection algorithms rely on deep networks which perform best in the presence of a large amount of training data. This work aims to demonstrate the utility of synthetic (both live and spoof) fingerprints in supplying these algorithms with sufficient data to improve the performance of fingerprint spoof detection algorithms beyond the capabilities when training on a limited amount of publicly available ``real'' datasets. First, we provide details of our approach in modifying a state-of-the-art generative architecture to synthesize high quality live and spoof fingerprints. Then, we provide quantitative and qualitative analysis to verify the quality of our synthetic fingerprints in mimicking the distribution of real data samples. We showcase the utility of our synthetic live and spoof fingerprints in training a deep network for fingerprint spoof detection, which dramatically boosts the performance across three different evaluation datasets compared to an identical model trained on real data alone. Finally, we demonstrate that only 25\% of the original (real) dataset is required to obtain similar detection performance when augmenting the training dataset with synthetic data.
\end{abstract}

% Note that keywords are not normally used for peerreview papers.
\begin{IEEEkeywords}
Generative Adversarial Networks, Fingerprint Presentation Attack Detection, Fingerprint Spoof Detection, Synthetic Fingerprint Generation
\end{IEEEkeywords}

\IEEEpeerreviewmaketitle

\section{Introduction}

\begin{figure}[t]
\includegraphics[width=\linewidth]{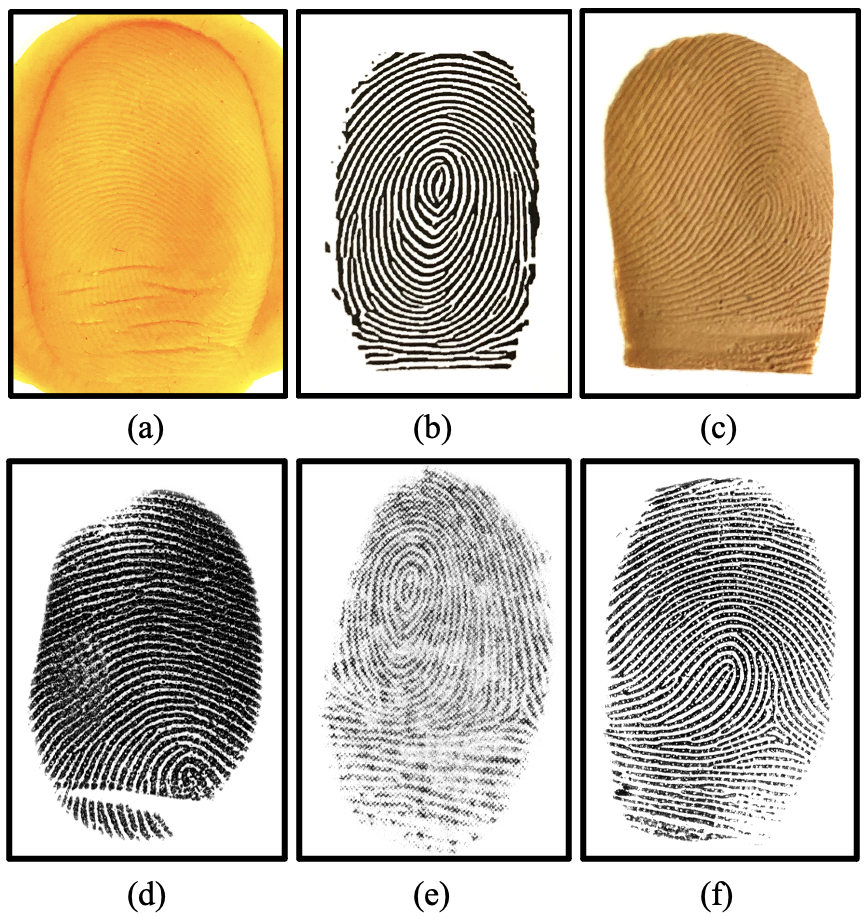} 
\caption{Example fabricated fingerprint spoofs of various materials and corresponding fingerprint impressions captured on a CrossMatch Guardian200 fingerprint reader. (a) PlayDoh spoof, (b) printed paper spoof, (c) latex spoof, (d) fingerprint impression from the PlayDoh spoof, (e) fingerprint impression from the printed paper spoof, and (f) fingerprint impression from the latex spoof.}
\label{fig:examples}
\end{figure}

\IEEEPARstart{F}{ingerprint} recognition has had a long history in person identification due to the purported uniqueness and permanence of fingerprints, originally pointed out by Sir Francis Galton in his 1982 book titled ``Finger Prints''~\cite{galton1892finger} and reaffirmed in many works over the last century; including the well-known studies on the individuality and longitudinal permanence of fingerprint recognition~\cite{pankanti2002individuality, yoon2015longitudinal}. Clearly, a significant contributor to their wide spread adoption is the high level of verification performance achieved by state-of-the-art (SOTA) algorithms for automated fingerprint recognition\footnote{Today's top performing algorithm in the FVCongoing 1:1 hard benchmark achieved a False Non-Match Rate (FNMR) of just 0.626\% at a False Match Rate (FMR) of 0.01\%~\cite{fvc_ongoing}}. However, despite the impressive accuracy achieved to date by the top performing fingerprint recognition algorithms, there remains many on-going efforts to further improve the capabilities of fingerprint recognition systems - especially in terms of recognition speed and system security. As a result, there has been a recent push toward deep neural network (DNN) based models for fingerprint recognition~\cite{cao2017fingerprint, song2017fingerprint, song2019aggregating, li2019learning, deepprint, lin2018cnn, grosz2021c2cl}. These compact, fixed-length embeddings can be matched efficiently and combined with homomorphic encryption for added security~\cite{engelsma2022hers}. For a more exhaustive account of existing deep learning approaches to fingerprint recognition and other biometric modalites, interested readers are encouraged to consult one of the many surveys on deep learning in biometrics (e.g., \cite{minaee2019biometrics, sundararajan2018deep}).

Indeed, this push toward DNN-based fingerprint recognition comes in the wake of the success demonstrated in the face recognition domain in applying DNN models to face recognition, which was aided by the availability of large-scale face recognition databases which were easily crawled from the web; despite the many ethical and privacy concerns which have led to many of these datasets to be recalled today. Arguably, at least in part, the reason for the delayed adoption of DNNs for fingerprint recognition has been the lack of publicly available, large-scale fingerprint recognition datasets and increased scrutiny over privacy of biometric data, which has led to many works to generate synthetic fingerprint images~\cite{engelsma2022printsgan, bouzaglo2022synthesis, bahmani2021high, level-3, mistry2019fingerprint, kai_synthetic, lightweight, synfi, attia2019fingerprint, finger-gan, cappelli2002synthetic, zhao2012fingerprint, johnson2013texture, bontrager2018deepmasterprints}.

Similarly, there has been an increased interest in DNN-based models for fingerprint spoof, i.e., presentation attack (PA), detection; where the scale and amount of publicly available data is also limited. Table~\ref{table:datasets} gives a list of the publicly available fingerprint spoof datasets. Compared to the largest, public fingerprint recognition dataset, NIST Special Database 302~\cite{nist302}, which contains fingerprints from 2,000 unique fingers, the largest publicly available fingerprint spoof datasets, e.g., the LivDet competition datasets, contain at most 1,000 unique fingers (for the Swipe sensor in LivDet 2013~\cite{ghiani2013livdet})\footnote{Swipe sensors are no longer in vogue after Apple introduced the "area capacitive sensor" in Touch ID.}. Compounding the problem is the difficulty in collecting large-scale fingerprint spoof datasets due to the increased time and complexity in fabricating and collecting spoof fingerprints. All of which motivates the potential of synthetic data as a viable alternative; however, to the best of our knowledge, there does not exist a synthetic fingerprint spoof generator to fill the gap between the amount of publicly available fingerprint spoof data and training of data-hungry deep learning based models.

To address the lack of large-scale fingerprint spoof datasets, we propose SpoofGAN. Inspired by the impressive results of the recently proposed PrintsGAN~\cite{engelsma2022printsgan}, SpoofGAN is a multi-stage generative architecture to fingerprint generation. SpoofGAN is different from PrintsGAN in the following ways:

\begin{itemize}
    \item Generation of plain print fingerprints, which compared to the rolled prints generated by PrintsGAN, are more representative of the publicly available spoof fingerprint datasets and exhibit different textural characteristics, distortions, etc.
    \item Ability to synthesize both live (i.e., bona fide) and spoof samples of the same fingerprint identity.
    \item Replacing the learned warping and cropping module with a statistical, controllable non-linear deformation model to synthesize multiple, realistic impressions per finger. This allows us to control the degree of distortion applied.
\end{itemize}

In this work, we use the terms \textit{live} and \textit{bona fide} interchangeably; however, live is not synonymous with real. We make the distinction that real fingerprint images are those captured on a fingerprint reader by either a real human finger or physical spoof artifact, whereas synthetic fingerprints are digital renderings of fingerprint like images. Thus, there can be both synthetic live fingerprint images and synthetic spoof fingerprint images that are generated, just as there are real live fingerprint images and real spoof fingerprint images.

We validate the realism of our synthetic live and spoof images through extensive qualitative and quantitative metrics including NFIQ2~\cite{nfiq}, minutiae statistics, match scores from a SOTA fingerprint matcher, and T-SNE feature space analysis showing the similarity of real live and spoof embeddings to the embeddings of our synthetic live and spoof fingerprints. Besides verifying the realism of our synthetic spoof generator, we also show how SpoofGAN fingerprints can be used to train a DNN for fingerprint spoof detection. We show this by improving the performance of a spoof detection model by augmenting an existing fingerprint spoof dataset with additional samples from our synthetic generator. We also open the door to jointly optimizing for fingerprint spoof detection and recognition in an end-to-end learning framework with our ability to generate a large-scale dataset of multiple impressions per finger of both live and spoof examples.

More concisely, the contributions of this research are as follows:
\begin{itemize}
    \item A highly realistic plain print synthetic fingerprint generator capable of generating multiple impressions per finger. 
    \item The first, to the best of our knowledge, synthetic fingerprint spoof generator which is capable of producing both live and spoof impressions of the same identity. This opens the door to joint optimization of fingerprint spoof detection and recognition algorithms.
    \item Quantitative and qualitative analysis to verify the quality of our generated live and spoof fingerprints.
    \item Experiments showcasing improved fingerprint spoof detection when augmenting existing fingerprint datasets with our synthetic live and spoof fingerprints.
\end{itemize}

\begin{table*}
\caption{Publicly available fingerprint spoof datasets.}
\centering 
\begin{threeparttable}
\begin{tabular}{|L{0.13\linewidth}C{0.09\linewidth}C{0.09\linewidth}C{0.48\linewidth}C{0.09\linewidth}|}
\noalign{\hrule height 1.0pt}
\specialcell{\textbf{Name}} & \textbf{\specialcell{\# Train Images \\ Live (Spoof)}} & \textbf{\specialcell{\# Test Images \\ Live (Spoof)}} & \specialcell{\textbf{Spoof types}} & \specialcell{\textbf{Sensors}}\\
\noalign{\hrule height 1.0pt}
LivDet 2009~\cite{marcialis2009first} & \begin{tabular}[c]{@{}c@{}}1000 (1000)\\ 1000 (1000)\end{tabular} & \begin{tabular}[c]{@{}c@{}}1000 (1000)\\ 1000 (1000)\end{tabular} & Ecoflex, Gelatine, Latex, Modasil, WoodGlue & \begin{tabular}[c]{@{}c@{}}Biometrika\\ Italdata\end{tabular}\\
\noalign{\hrule height 0.5pt}
LivDet 2011~\cite{yambay2012livdet} & \begin{tabular}[c]{@{}c@{}}1000 (1000)\\ 1000 (1000)\\ 1000 (1000)\\ 1000 (1000)\end{tabular} & \begin{tabular}[c]{@{}c@{}}1000 (1000)\\ 1000 (1000)\\ 1000 (1000)\\ 1000 (1000)\end{tabular} & Gelatine, latex, PlayDoh, Silicone, Wood Glue, Ecoflex & \begin{tabular}[c]{@{}c@{}}Biometrika\\ DigitalPersona\\ ItalData\\ Sagem\end{tabular}\\
\noalign{\hrule height 0.5pt}
LivDet 2013~\cite{ghiani2013livdet} & \begin{tabular}[c]{@{}c@{}}1000 (1000)\\ 1000 (1000)\\ 1250 (1000)\\ 1250 (1000)\end{tabular} & \begin{tabular}[c]{@{}c@{}}1000 (1000)\\ 1000 (1000)\\ 1250 (1000)\\ 1250 (1000)\end{tabular} & Body Double, Latex, PlayDoh, Wood Glue, Gelatine, Ecoflex, Modasil & \begin{tabular}[c]{@{}c@{}}Biometrika\\ ItalData\\ CrossMatch\\ Swipe\end{tabular}\\
\noalign{\hrule height 0.5pt}
LivDet 2015~\cite{mura2015livdet} & \begin{tabular}[c]{@{}c@{}}1000 (1000)\\ 1000 (1000)\\ 1000 (1000)\\ 1510 (1473)\end{tabular} & \begin{tabular}[c]{@{}c@{}}1000 (1500)\\ 1000 (1500)\\ 1000 (1500)\\ 1500 (1448)\end{tabular} & Ecoflex, Gelatine, Latex, Liquid Ecoflex, RTV, WoodGlue, Body Double, PlayDoh, OOMOO & \begin{tabular}[c]{@{}c@{}}Biometrika\\ DigitalPersona\\ GreenBit\\ CrossMatch\end{tabular}\\
\noalign{\hrule height 0.5pt}
LivDet 2017~\cite{mura2018livdet}    & \begin{tabular}[c]{@{}c@{}}1000 (1200)\\ 1000 (1200)\\ 999 (1199)\end{tabular} & \begin{tabular}[c]{@{}c@{}}1700 (2040)\\ 1700 (2676)\\ 1700 (2028)\end{tabular} & Body Double, Ecoflex, Wood Glue, Gelatine, Latex, Liquid Ecoflex & \begin{tabular}[c]{@{}c@{}}GreenBit\\ Orcanthus\\ DigitalPersona\end{tabular}\\
\noalign{\hrule height 0.5pt}
LivDet 2019~\cite{livdet2019}    & \begin{tabular}[c]{@{}c@{}}1000 (1200)\\ 1000 (1200)\\ 1000 (1000)\end{tabular} & \begin{tabular}[c]{@{}c@{}}1020 (1224)\\ 990 (1088)\\ 1019 (1224)\end{tabular} & Body Double, Ecoflex, Wood Glue, Gelatine, Latex, Liquid Ecoflex & \begin{tabular}[c]{@{}c@{}}GreenBit\\ Orcanthus\\ DigitalPersona\end{tabular}\\
\noalign{\hrule height 0.5pt}
LivDet 2021~\cite{casula2021livdet}    & \begin{tabular}[c]{@{}c@{}}1250 (1500)\\ 1250 (1500)\end{tabular} & \begin{tabular}[c]{@{}c@{}}2050 (2460)\\ 2050 (2460)\end{tabular} & Latex, RProFast, Nex Mix 1, Body Double, Elmer's Glue, GLS20, RFast30 & \begin{tabular}[c]{@{}c@{}}GreenBit\\ Dermalog\end{tabular}\\
\noalign{\hrule height 0.5pt}
MSU FPAD~\cite{chugh2018fingerprint} & \begin{tabular}[c]{@{}c@{}}2250 (3000)\\ 2250 (2250)\end{tabular} & \begin{tabular}[c]{@{}c@{}}2250 (3000)\\ 2250 (2250)\end{tabular} & Ecoflex, PlayDoh, 2D Matte Paper, 2D Transparency & CrossMatch\\
\noalign{\hrule height 0.5pt}
MSU FPAD v2~\cite{chugh2019fingerprint} & 4743 (4912) & 1000 (leave-one-out) & 2D Printed Paper, 3D Universal Targets, Conductive Ink on Paper, Dragon Skin, Gelatine, Gold Fingers, Latex Body Paint, Monster Liquid Latex, Play Doh, Silicone, Transparency, Wood Glue & CrossMatch\\
\noalign{\hrule height 1.0pt}
\end{tabular}
\begin{tablenotes}
\item[1] The dataset release agreement for all LivDet databases can be found at \url{https://livdet.org/registration.php}.
\item[2] Similarly, the dataset release form for the MSU FPAD dataset can be found at \url{http://biometrics.cse.msu.edu/Publications/Databases/MSU_FPAD/}
\end{tablenotes}
\end{threeparttable}
\label{table:datasets}
\end{table*}

\section{Related Work}

\subsection{Fingerprint Spoof Detection}
One significant risk to the security of fingerprint recognition systems is that of presentation attacks, defined by ISO standard IEC 30107-1:2016(E) as a “presentation to the biometric data capture subsystem with the goal of interfering with the operation of the biometric system”~\cite{isopad}. The most common type of presentation attacks are spoof attacks, i.e., physical representations of finger-like structures aimed at either mimicking the fingerprint ridge-valley structure of another individual or subverting the user's own identity. Spoof attacks may come in many different forms and materials such as those shown in Figure~\ref{fig:examples}. 

Several hardware and software-based solutions to detecting spoof attacks have been proposed. Hardware based solutions include specialized sensors that leverage various ``liveness'' cues at the time of acquisition, such as conductivity of the material/finger, sub-dermal imaging, and multi-spectral lighting~\cite{baldisserra2006fake, lapsley1998anti, engelsma2018raspireader,chugh2019oct,tolosana2018towards,spinoulas2021multi}. On the other hand, software based solutions typically rely on only the information captured in the grayscale image acquired by the fingerprint reader~\cite{marcialis2010analysis, marasco2012combining, ghiani2012fingerprint, ghiani2013fingerprint, nogueira2016fingerprint, pala2017deep, chugh2018fingerprint}. Despite the limited publicly available fingerprint spoof data, many of the state-of-the-art software-based solutions to fingerprint spoof detection leverage convolutional neural networks to learn the decision boundary between live and spoof images. Some researchers have proposed training their algorithms on smaller patches of the fingerprint images as a way to deal with limited amounts of available training data, which roughly increases the number of training images by a factor proportional to the number of patches~\cite{chugh2018fingerprint}. However, given the increased scrutiny over privacy concerns related to biometric datasets, it is not certain whether any spoof fingerprint datasets will remain available in the future. Thus, motivating the need for synthetic data to fill this gap.

Another challenge related to limited training data is that of unseen presentation attacks, or fingerprint images arising from never before seen PA instruments. This problem is also commonly referred to in the literature as cross-material generalization. Some strategies proposed to improve the cross-material performance of spoof detection include learning a tighter boundary around the live class via one-class classifiers~\cite{engelsma2019generalizing, ding2016ensemble}, incorporating adversarial representational learning to encourage robustness to varying material types~\cite{grosz2020fingerprint, pereira2020robust}, or applying style transfer to mix textures from some known PA materials to better fill the space of unknown texture characteristics that may be encountered~\cite{chugh2020fingerprint, gajawada2019universal}. Similar ideas may apply to synthetic data generation, where new material types can be synthesized by mixing characteristics of known PAs.

\begin{figure*}[t]
\includegraphics[width=1.0\linewidth]{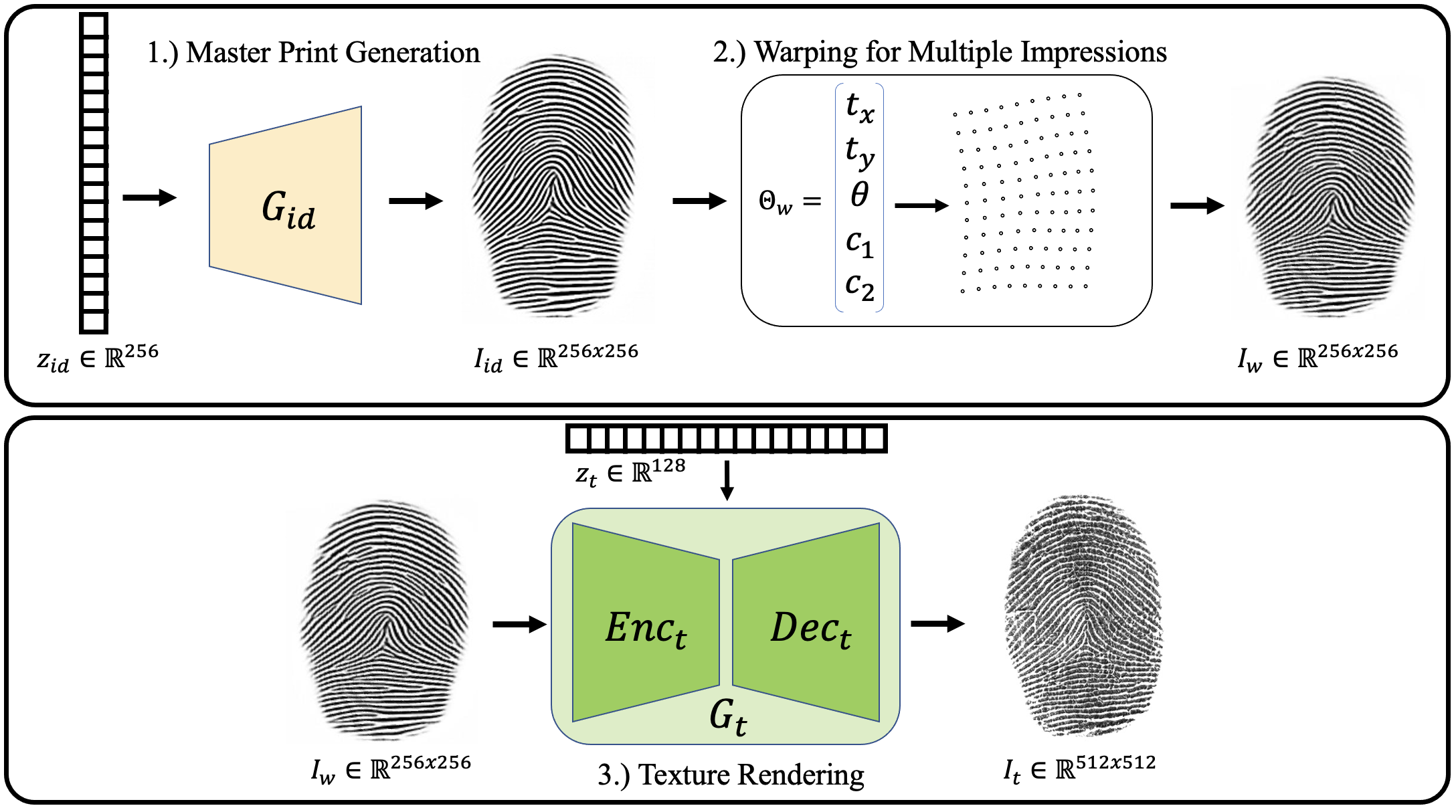} 
\caption{Overview of the SpoofGAN Architecture.}
\label{fig:schematic}
\end{figure*}

\section{Proposed Synthetic Spoof Fingerprint Generator}
In this section, we detail the process of generating synthetic live and spoof fingerprints. Motivated by the success of previous multi-stage fingerprint generation methods (e.g., \cite{engelsma2022printsgan, cappelli2002synthetic, wyzykowski2021level}), SpoofGAN generates highly realistic fingerprints in multiple stages. First, unique fingerprint identities are synthesized through generating binary master fingerprints which define the fingerprint ridge structure. Following the master print synthesis stage, perturbations such as random rotation, translation, and non-linear deformation are applied to simulate realistic, repeat impressions. Finally, each generated fingerprint impression is input to a second neural network to impart realistic textures which mimic a database of real fingerprints. An overview of the entire process is given in Figure~\ref{fig:schematic}.

\subsection{Master Print Synthesis}
The first step in generating synthetic fingerprints with SpoofGAN is generating binary master fingerprints $I_{id}\in~\{0, 1\}^{256\times256}$ from a random vector $z_{id}\in\mathcal{R}^{256}$ sampled from a standard normal distribution (i.e., $z_{id} \sim \mathcal{N}(0,1)$). In particular, we used a standard BigGAN~\cite{brock2018large} architecture for this task, consisting of a generator $G_{id}$ and a discriminator $D_{id}$. Since many spoof impressions can exhibit non-realistic fingerprint ridge structures, either from artifacts introduced in the fabrication (e.g., bubbles in the ridges) or in the presentation process (e.g., smudges due to the high elasticity of some spoof material types), we chose to train $G_{id}$ using a database of only live fingerprint impressions consisting of 38,164 images captured on a CrossMatch Guardian200 fingerprint reader. As we will show later, these artifacts for the spoof impressions can be introduced in the later texture rendering stage of our synthesis pipeline. The network is trained via an adversarial loss shown in Equation~\ref{eq:adv_1}, where $I$ is a binary fingerprint image extracted from a real fingerprint image using the Verifinger v12.0 SDK.

\begin{equation}
    \mathcal{L}_{adv}(G, D) =  \mathbb{E}_{I}~[logD(I)] + \mathbb{E}_z~[log(1-D(G(z)))]
    \label{eq:adv_1}
\end{equation}

\subsection{Generating Multiple Realistic Impressions}
To generate multiple impressions from a single master print, we apply realistic rotation, translation, and non-linear deformation for each subsequent impression. Rotations are applied via a uniform random sampling in the range [-30\degree, 30\degree], whereas translations in both the x and y directions are uniformly sampled in the range [-25, 25] pixels. Finally, realistic non-linear deformations are applied via a learned, statistical deformation model proposed by Si et al. in \cite{si2015detection}. The distortion parameters were learned from a database of 320 distorted fingerprint videos in which the minutiae locations in the first and last frames were manually labeled and the displacements between corresponding minutiae points were used to estimate a distortion field via a Thin-Plate-Spline deformation model~\cite{bookstein1989principal}. The distortion fields were condensed into a subset of eigenvectors, $e_i$, computed from a Principal Component Analysis of the covariance matrix estimated from the 320 videos. By varying the coefficients, $c_i$, multiplied to each of the eigenvectors, we vary the magnitude of distortion, $d$, applied to an input fingerprint along realistic distortion directions according to the equation\ref{eq:distortion}, where $\lambda_i$ are the eigenvalues of each eigenvector and $\bar{d}$ is the average distortion field. For our implementation, we randomly sample the coefficients of the two largest eigenvectors from a normal distribution with mean 0 and standard deviation of 0.66, which were empirically determined to produce reasonable distortions.

\begin{equation}
    d \approx \bar{d}~\sum_{i = 1}^{t}c_i\sqrt{\lambda_ie_i}
    \label{eq:distortion}
\end{equation}

\subsection{Texture Rendering}
The final stage of our fingerprint generation process consists of imparting each fingerprint with a realistic texture that mimics the distribution of real live and spoof images. For the generator, $G_t$, we use an encoder-decoder architecture which translates an input, warped binary image, $I_w$, into a realistic fingerprint impression, $I_t$. To promote diversity in the rendered images, a random texture vector sampled from a standard normal distribution (i.e., $z_t \sim \mathcal{N}(0,1)$) is injected into the network and encoded into $\gamma$ and $\beta$ parameters for performing instance normalization on the intermediate feature maps of $G_t$. Finally, the discriminator, $D_t$ utilizes the same architecture used in the binary master print synthesis network.

The goal of our texture renderer is two fold: i.) generate realistic texture details and ii.) maintain the identity of the rendered fingerprint between corresponding impressions of the same finger. Thus, we introduce two losses in addition to the conventional GAN loss (eq.~\ref{eq:gan_loss_T}) to maintain the identity of textured fingerprints. The first is an identity loss to minimize the $L_2$ distance between feature embeddings of corresponding fingerprint impressions using a SOTA fingerprint matcher DeepPrint~\cite{deepprint}  (eq.~\ref{eq:dp_loss_T}) and the other is an $L_2$ pixel loss between ground truth binary images and binary images extracted from the textured fingerprints (eg.~\ref{eq:image_loss_T}). The $L_2$ pixel loss is computed on the binary images, rather than the grayscale images, to allow for the network to generate diverse ``styles'' in the generated fingerprints to simulate different pressure, moisture content, and contrast in subsequent impressions; all of which would lead to slightly different loss values compared to the ground truth image unless first converted to binary ridge images. To make the process of binarization of the generated fingerprints differentiable, we train a convolutional autoencoder to binarize input fingerprints which is trained on 38,164 grayscale/binary image pairs. The overall losses for $G_t$ and $D_t$ are given in equations \ref{eq:overall_loss_G_T} and \ref{eq:overall_loss_D_T}, respectively.

\begin{enumerate}
    \item GAN loss: Classical min-max GAN loss between the discriminator, $D_t(\cdot)$, trying to classify each original fingerprint image, $I$, as real and each synthetic fingerprint $I_t=G_t(I_w)$ as fake. Meanwhile, $G_t(\cdot)$ is trying to fool $D_t(\cdot)$ into thinking its outputs come from the original image distribution.
        \begin{equation}
        \label{eq:gan_loss_T}
        \mathcal{L}_{adv} =  \mathbb{E}_I~[logD(I)] + \mathbb{E}_{I_w}~[log(1-D(G(I_w)))]
        \end{equation}
    \item DeepPrint loss: $L_2$ distance between the DeepPrint embedding, $R$, extracted from the ground truth grayscale image and the DeepPrint embedding, $\hat{R}$, extracted from the synthesized grayscale fingerprint image. 
        \begin{equation}
        \label{eq:dp_loss_T}
        L_{dp} = \frac{1}{2}\sum(R - \hat{R})^2 \\
        \end{equation}
    \item Image/pixel loss: $L_2$ loss between the ground truth binary fingerprint image, $I_w$, and synthesized binary fingerprint image, $\hat{I_w}$.
        \begin{equation}
        \label{eq:image_loss_T}
        L_i = \frac{1}{2}\sum_{x,y} (I_w(x,y) - \hat{I_w}(x,y))^2
        \end{equation}
    \item Overall loss for $G_{t}(\cdot)$: $\lambda_{1}=1$, $\lambda_{2}=2$, and $\lambda_{3}=10$ (determined empirically). 
        \begin{equation}
        \label{eq:overall_loss_G_T}
        \mathcal{L}_{G_t} = \lambda_1\mathcal{L}_{adv} + \lambda_2\mathcal{L}_{dp} + \lambda_3\mathcal{L}_{i}
        \end{equation}
    \item Overall loss for $D_{t}(\cdot)$:
        \begin{equation}
        \label{eq:overall_loss_D_T}
        \mathcal{L}_{D_t} = \mathcal{L}_{adv}
        \end{equation}
\end{enumerate}

Unlike the binary master print synthesis and warping stages, an individual texture rendering network is trained for each material type (live, ecoflex spoof, PlayDoh spoof, etc.). Due to the limited number of images in our spoof dataset, we pretrained a texture rendering network on the 282K unique fingerprint database taken from the MSP longitudinal database introduced in~\cite{yoon2015longitudinal}\footnote{This database is not publicly available, but the pretrained model can be made available upon request.}. Initially, following the pretraining, two texture rendering networks are trained further, one on the dataset of 38,164 live only impressions and the other on the 3,366 spoof fingerprint images consisting of all spoof types aggregated together. Finally, we further finetune the model trained on all spoofs for each of the individual spoof types to give more fine-grained control on the specific spoof style being generated.

Due to the limited number of images in our spoof dataset, we pretrained a texture rendering network on the 282K unique fingerprint database taken from the MSP longitudinal database introduced in~\cite{yoon2015longitudinal}\footnote{This database is not publicly available, but the pretrained model can be made available upon request.}. Following the pretraining, two texture rendering networks are finetuned; one on the dataset of 38,164 live only impressions and the other on the 3,366 spoof fingerprint images consisting of all spoof types aggregated together. Finally, we further finetune the model trained on all spoofs for each of the individual spoof types to give more fine-grained control on the specific spoof style being generated. Thus, unlike the binary master print synthesis and warping stages which are shared, each individual spoof type has it's own rendering network. Alternatively, a conditional GAN structure could be used to generate spoof classes of each type within a single network; however, we found that due to the very limited amount of training images in some spoof types (e.g., 50 images), finetuning for just a few epochs on each spoof type individually produced higher quality images.

\section{Experimental Results}
In this section, we aim to validate the realism of our synthetic live and spoof images via several qualitative and quantitative experiments. First, we provide details on the datasets involved in the following experiments, followed by some example fingerprint images generated by SpoofGAN to qualitatively compare with real fingerprint images. Finally, several quantitative metrics are used to compare the utility and distribution of SpoofGAN generated fingerprint images compared to the real fingerprint images.

\bgroup
\def\arraystretch{1.5}
\begin{table*}
\caption{Summary of the spoof datasets used in our experiments.}
\begin{tabularx}{\textwidth}{c||c||c||c||c}
\noalign{\hrule height 1.5pt}
\textbf{Dataset}            & \textbf{LivDet 2013}                   & \textbf{LivDet 2015}                           & \textbf{GCT 1-5}                                                                                          & \textbf{GCT 6}                                          \\
\hline
\noalign{\hrule height 1.0pt}
Fingerprint Reader          & CrossMatch                             & CrossMatch                                     & CrossMatch                                                                                                & CrossMatch                                              \\
\hline
Model                       & L Scan Guardian                        & L Scan Guardian                                & Guardian200                                                                                               & Guardian200                                             \\
\hline
Resolution (dpi)            & 500                                    & 500                                            & 500                                                                                                       & 500                                                     \\
\hline
\specialcell{\# Live Images \\ (Train/Test)} & 1250 / 1250                            & 1510 / 1500                                    & 38,164 / 0                                                                                                & 7,357 / 14,236                                          \\
\hline
\specialcell{\# PA Images \\ (Train/Test)}   & 500 / 440                              & 1473 / 1448                                    & 3,366 / 0                                                                                                 & 2,550 / 1,829                                           \\
\hline
PA Materials                & \specialcell{Body Double, Latex, \\ PlayDoh, Wood Glue} & \specialcell{Ecoflex, Gelatine, Body \\ Double, PlayDoh, OOMOO} & \specialcell{Dragon Skin, Ecoflex, Paper, Silicone, \\ Transparency, Gelatine, Glue, PDMS, \\ Knox Gelatine, Gummy Overlay, Tattoo} & \specialcell{Ecoflex, Silicone, Gummy \\ Overlay, Tattoo, Knox Gelatine}\\
\noalign{\hrule height 1.5pt}
\end{tabularx}
\label{tab:dataset_stats}
\end{table*}
\egroup

\subsection{Datasets}
A main motivation for this paper is the lack of large-scale, publicly available fingerprint spoof datasets. Some of the largest datasets that are available have resulted from the annual LivDet competition series dating as far back as 2009~\cite{yambay2012livdet,ghiani2013livdet, mura2015livdet, mura2018livdet, livdet2019, casula2021livdet}. A more comprehensive list of the fingerprint spoof datasets currently available to the research community is given in Table~\ref{table:datasets}, whereas the datasets used in this paper are given in Table~\ref{tab:dataset_stats}. In this paper we focus our experiments on fingerprint images obtained via the CrossMatch optical reader from LivDet 2013, LivDet 2015, and the Government Controlled Test (GCT) dataset of live and spoof fingerprints collected as part of the IARPA ODIN program\footnote{We selected CrossMatch for our experiments since it is one of the most popular slap (4-4-2) capture readers used in law enforcement, homeland security and civil registry applications.}. Our training dataset for SpoofGAN, referred to as GCT 1-5, consists of 38,164 live fingerprint images and 3,366 spoof fingerprint images from 2,007 fingers and 11 different spoof types (Dragon Skin, Ecoflex, Paper, Silicone, Transparency, Gelatine, Glue, PDMS, Knox Gelatine, Gummy Overlay, and Tattoo). For our evaluations, we have followed the same train/test protocol referenced in LivDet2015 and LivDet2013, as well as reserved a fraction of the GCT dataset, GCT 6, as an evaluation dataset.

\subsection{Qualitative Analysis of Synthetic Live and Spoof Images}
\begin{figure*}
\centering
\includegraphics[height=0.94\linewidth,angle=270,origin=c]{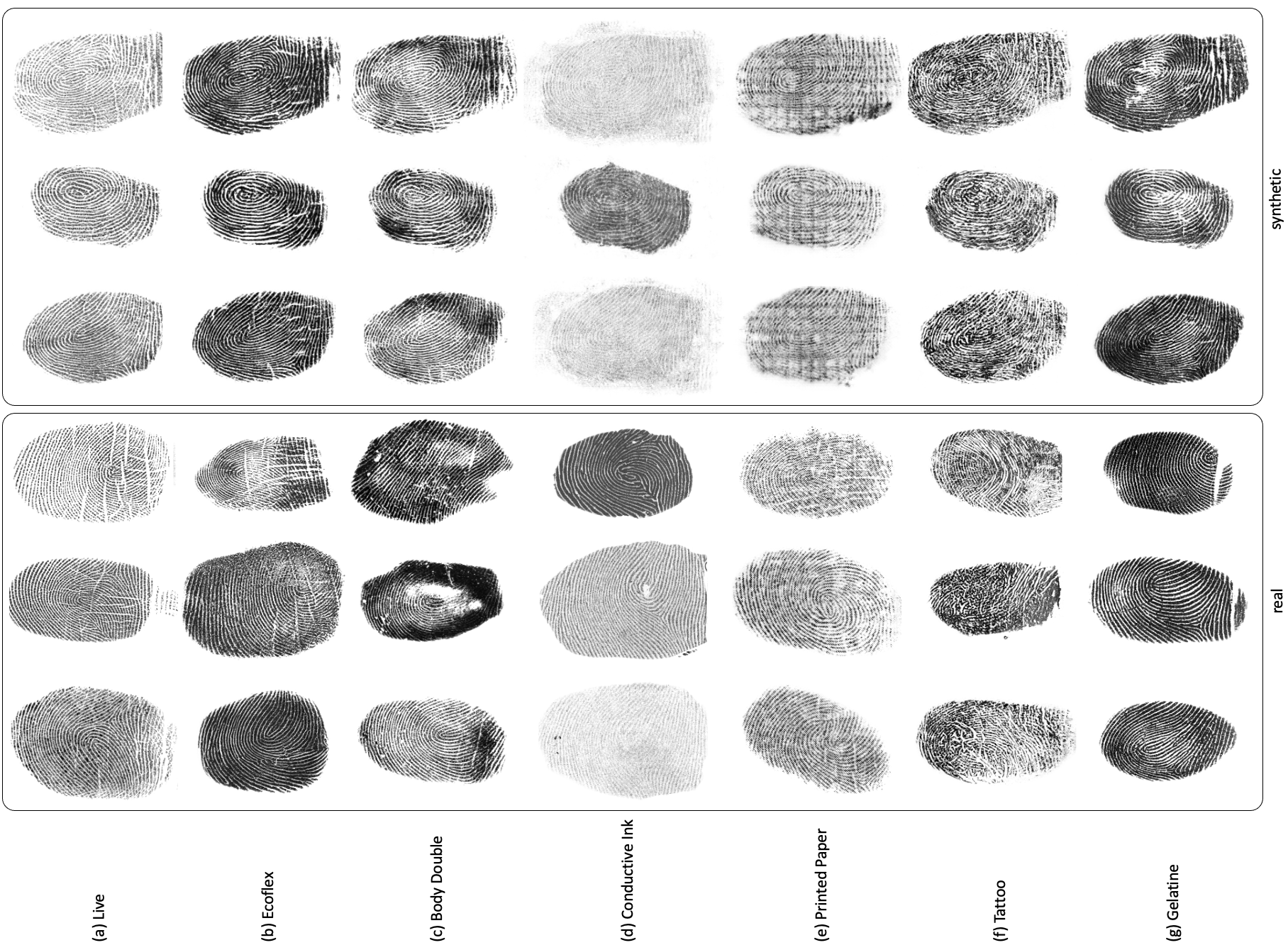}
\caption{Example real live and spoof images and synthetic live and spoof images generated by SpoofGAN of various material types: (a) live, (b) ecoflex, (c) body double, (d) conductive ink on paper, (e) 2d printed paper, (f) tattoo, and (g) gelatine.}
\label{fig:examples_rendered}
\end{figure*}

\begin{figure*}
\centering
\includegraphics[width=1.0\linewidth]{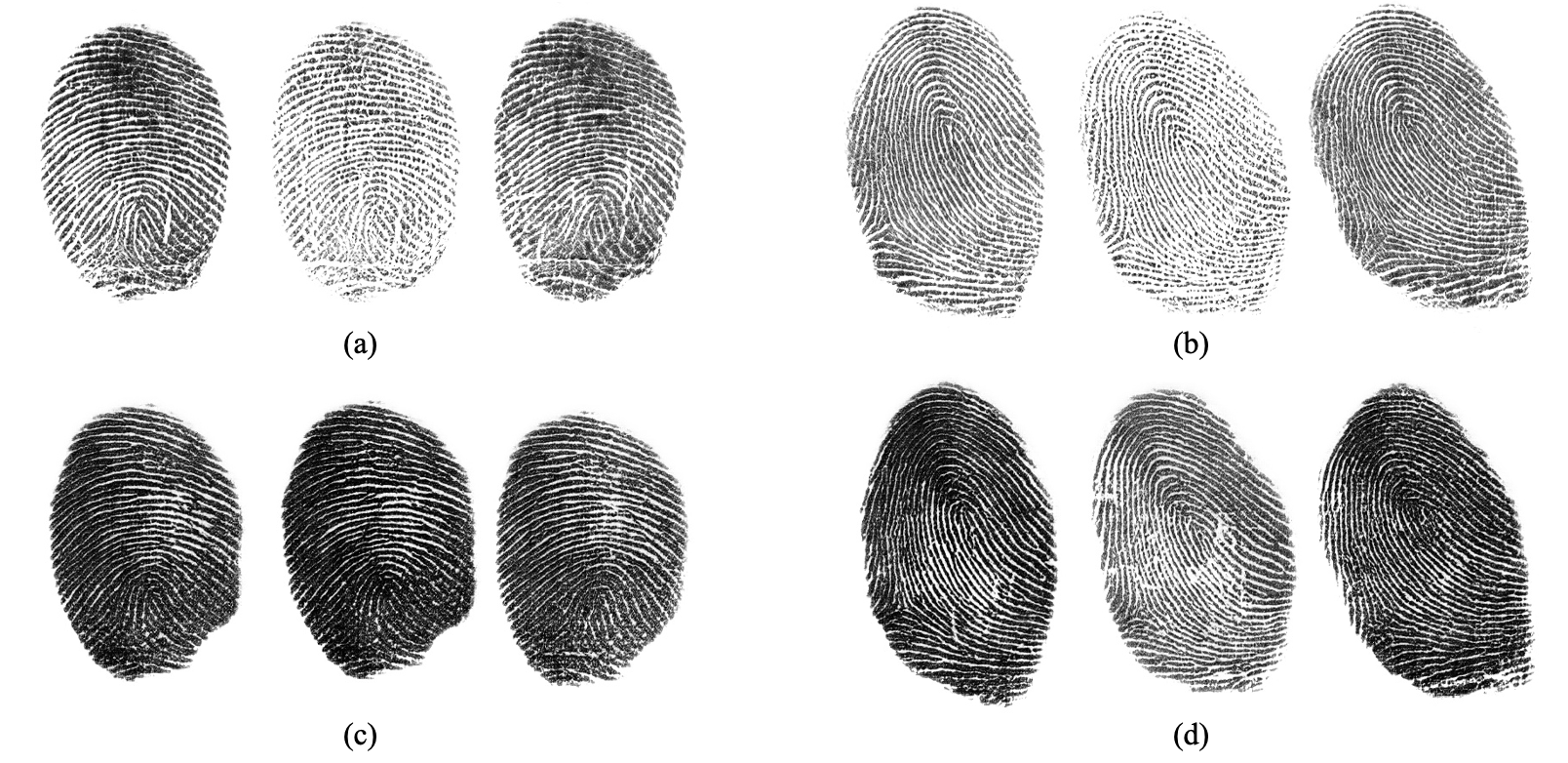}
\caption{Example images of multiple impressions of the same finger generated by SpoofGAN. (a) and (b) show three impressions each of two fingers rendered in a live style, whereas (c) and (d) show three impressions each of the same two fingers in a spoof style (ecoflex and body double, respectively).}
\label{fig:examples_rendered_longitudinal}
\end{figure*}

Example synthetic live and spoof fingerprints of varying material types (i.e., presentation Attack Instruments) are shown in column (b) of Figure~\ref{fig:examples_rendered} with corresponding real examples in column (a) shown for reference. Visually, SpoofGAN is generating spoof examples that closely resemble the target material type, which can be seen in the different texture characteristics for each material seen in the synthetic examples (e.g., Paper vs. Tattoo, etc.). Additionally, looking across the rows of the synthetic SpoofGAN images, we notice that the underlying identity is successfully being preserved across the different spoof styles being generated. Figure~\ref{fig:examples_rendered_longitudinal} further highlights SpoofGAN's ability to successfully generate multiple impressions of each finger in both live and spoof styles.

\subsection{Quantitative Analysis of Synthetic Live and Spoof Images}
Similar to previous works on synthetic fingerprint generation~\cite{bahmani2021high, engelsma2022printsgan}, we have evaluated the quality of SpoofGAN fingerprints on several quantitative metrics, including spoof detection performance by a pretrained spoof detector, the distribution of minutiae count, type, and quality extracted from SpoofGAN generated fingerprints compared to a real fingerprint dataset, NFIQ2 quality scores, match score distributions from a SOTA fingeprrint matcher, and identity leakage experiments.

\subsubsection{Spoof Detection Performance of Real vs. Synthetic Fingerprints}
\begin{figure}
\includegraphics[width=1.0\linewidth]{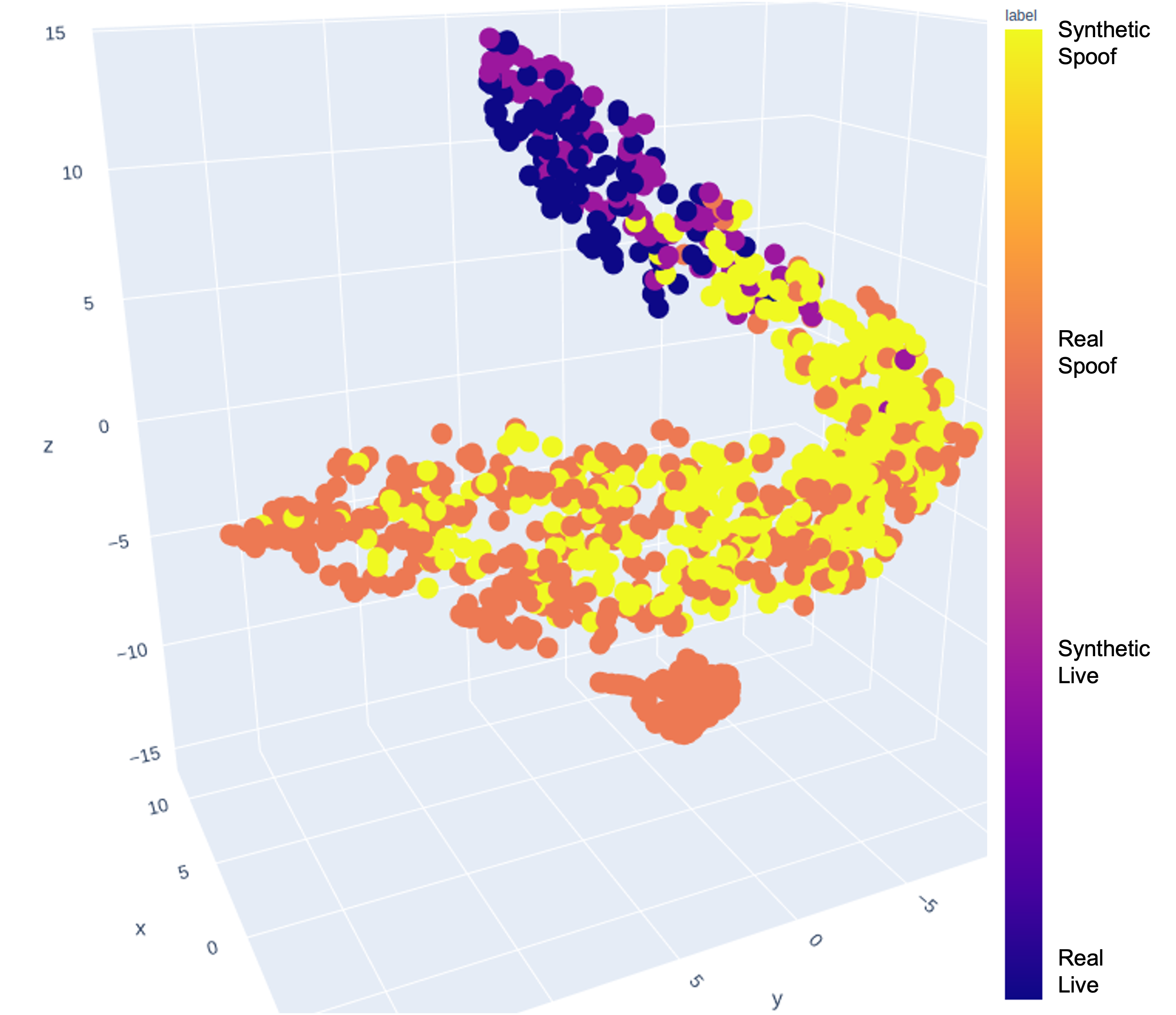} 
\caption{3D visualization of 2,048-dimensional embeddings of real live and spoof images from the LivDet2015 CrossMatch dataset compared with embeddings of our synthetic live and spoof images generated from SpoofGAN. Best viewed in color.}
\label{fig:tsne}
\end{figure}

Our first evaluation to verify the quality of our synthetic live and spoof fingerprints is to see whether a pretrained spoof detection algorithm trained on similar, real fingerprints performs equally well on our synthetic fingerprints. In particular, we pretrained an Inception v3 network on the GCT 1-5 data to classify between live and spoof fingerprint samples. Then, we evaluated the spoof detection performance on LivDet 2015 CrossMatch images compared to an equivalent sized database of synthetic fingerprints. As shown in Table~\ref{tab:pretrained_results}, the true detection rate (TDR) at a false detection rate (FDR) of 0.2\% is similar across multiple spoof types of both datasets, supporting our hypothesis that the synthetic samples should be useful in training additional spoof detection models without access to a large database of real live and spoof fingerprints for training\footnote{This metric is recommended by the IARPA ODIN program, one of the largest US government funded programs on spoof detection: \url{https://www.iarpa.gov/research-programs/odin}.}. Lastly, the embeddings of both real and synthetic images in the T-SNE embedding space suggest high similarity between the embeddings of real and synthetic images (see Figure~\ref{fig:tsne}). 

\begin{table}[t]
\begin{threeparttable}
\caption{Spoof detection model trained on real fingerprints (GCT 1-5) and evaluated on LivDet 2015 CrossMatch images (top row) and an equivalent sized synthetic fingerprint dataset (bottom row)\tnote{1}. Results given in TDR @ FDR = 0.2\%.}
 \centering
\begin{tabular}{lccccc}
 \toprule
  & \specialcell{BodyDouble} & \specialcell{Ecoflex} & \specialcell{PlayDoh} & \specialcell{OOMOO} & \specialcell{Gelatine}\\
 \midrule
 \specialcell{Real} & 100\%      & 99.57\% & 98.58\% & 99.58\% & 100\% \\
 \midrule
 \specialcell{Synthetic} & 100\%      & 100\%   & 100\%   & 99.58\% & 99.12\% \\
 \bottomrule
\end{tabular}
\begin{tablenotes}
\item[1] There are 1,500 live and 1,448 spoof fingerprint test images for CrossMatch in LivDet 2015, which we have replicated with synthetic data. Specifically, the spoof images consist of 300 Body Double, 270 Ecoflex, 297 OOMOO, 281 PlayDoh, and 300 Gelatine images.
\end{tablenotes}
\label{tab:pretrained_results}
\end{threeparttable}
\end{table}

\subsubsection{Feature Similarity Between Real and Synthetic Fingerprints}
For synthetic fingerprint images to be useful as a substitute for real fingerprint images, the features between a database of real fingerprint images and synthetic fingerprint images should closely align. For this analysis, we computed several statistics from the LivDet 2015 CrossMatch training dataset (lives only) and 1,500 SpoofGAN live fingerprint images which are shown in Table~\ref{tab:fp_stats}. In terms of fingerprint area, SpoofGAN images are, on average, smaller compared to the real fingerprint database. Since our training dataset consists of images of all 10 fingers, there is a bias toward smaller fingerprints considering the thumb as a minority class. Given this assumption, it is perhaps unsurprising that a GAN-based generation approach might exaggerate this class imbalance and generate smaller fingerprint area impressions. This problem is related to mode-collapse and has been noted in several GAN related works~\cite{goodfellow2014generative,srivastava2017veegan,roth2017stabilizing}, with some recent papers proposing strategies to improve the generation process in class-imbalanced datasets~\cite{mariani2018bagan,huang2021enhanced}.

Next, we computed the NFIQ 2.0 quality metric~\cite{nfiq} on both datasets. The NFIQ 2.0 scores for SpoofGAN are, on average, lower compared to LivDet. However, since one of the features considered in NFIQ 2.0 is the fingerprint area, we recomputed the scores on a $256\times256$ center crop of each of the fingerprints and observed that independent of fingerprint area, the NFIQ scores between SpoofGAN images and LivDet are much more aligned ($36.88\pm10.18$ vs. $43.65\pm14.12$).

Lastly, we computed some additional metrics specific to the distribution of minutiae since many of the state-of-the-art fingerprint algorithms incorporate minutiae information. The average minutiae count and minutiae quality computed by Verifinger 12.0 are given in Table~\ref{tab:fp_stats}. The average number of minutiae found in SpoofGAN seems to be lower compared to the CrossMatch images from LivDet 2015; however, the minutiae per Megapixel is similar for both datasets (59.74 vs. 59.49). The minutiae quality given by Verifinger is also very similar between the two datasets (71.89 vs. 70.78). 

\begin{table*}[t!]
\centering
{\normalsize
\caption{\normalsize{Metrics for real and SpoofGAN fingerprint images. Minutiae quality and NFIQ2 scores have a range of [0, 100].}}
\begin{tabularx}{\textwidth}{l|YY|YY}
\noalign{\hrule height 1.5pt}
                                  & \multicolumn{2}{c|}{LivDet 2015 CrossMatch}                          & \multicolumn{2}{c}{SpoofGAN}                     \\
Measure                           & \multicolumn{1}{c}{Mean} & \multicolumn{1}{c|}{Std. Dev.} & \multicolumn{1}{c}{Mean} & \multicolumn{1}{c}{Std. Dev.} \\
\noalign{\hrule height 1.0pt}
Total Minutiae Count              & 55.56                    & 18.43                         & 40.45                    & 11.57                         \\
Ridge Ending Minutiae Count       & 30.58                    & 12.12                         & 20.99                    & 6.67                         \\
Ridge Bifurcation Minutiae Count  & 24.98                    & 9.18                         & 19.46                    & 6.61                          \\
Verifinger Minutiae Quality  & 71.89                    & 16.00                         & 70.78                    & 15.15                         \\
Fingerprint Area (Megapixels) & 0.93               & 0.30                      & 0.68               & 0.15                      \\
Fingerprint Image Quality (NFIQ2)                             & 60.18                    & 19.35                         & 44.34                    & 14.38    \\  
\noalign{\hrule height 1.5pt}
\end{tabularx}
\label{tab:fp_stats}}
\end{table*}

\subsubsection{Diversity in Generated Fingerprint Identities}
\begin{figure*}
\centering
\includegraphics[height=0.37\linewidth]{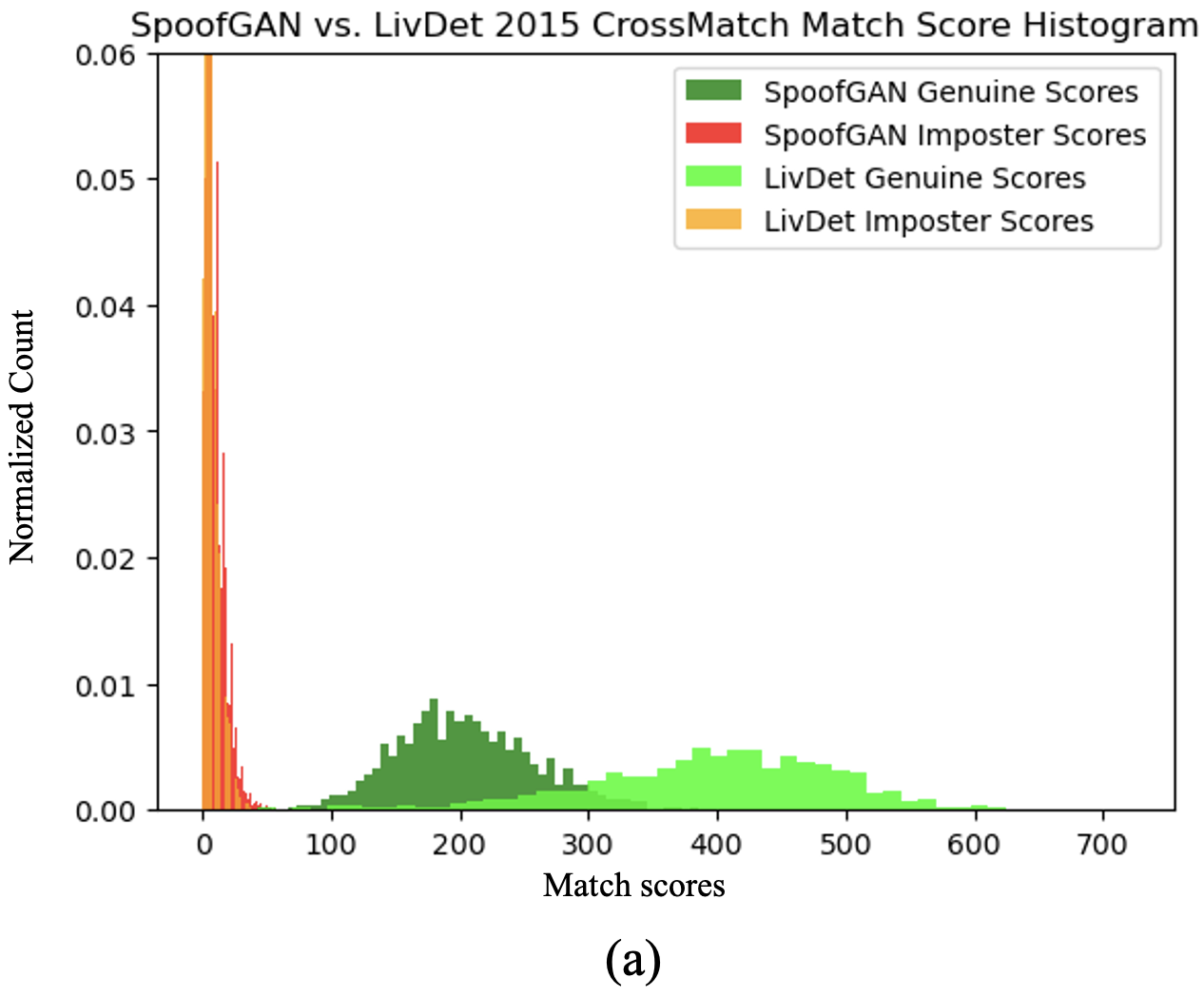}
\includegraphics[height=0.37\linewidth]{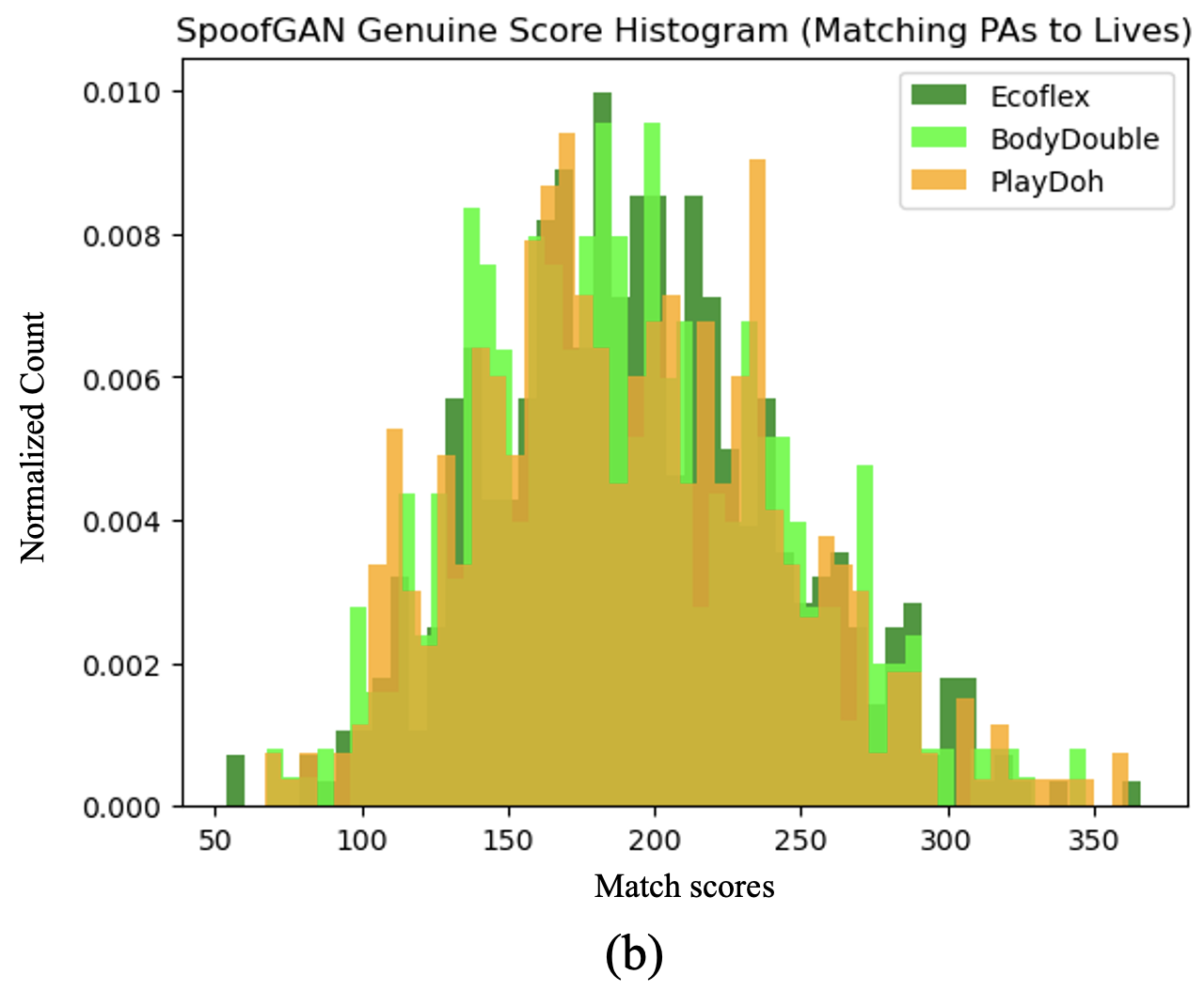}
\caption{Verifinger 12.0 match score distributions of the real LivDet 2015 CrossMatch L Scan Guardian database vs. an equivalent sized synthetic SpoofGAN database. (a) match scores computed between the live impressions of each dataset and (b) match scores computed between live and spoof impressions for the SpoofGAN images.}
\label{fig:verifinger_hists}
\end{figure*}

To verify that SpoofGAN generated fingerprints mimic the similarity score distribution of real live and spoof fingerprints, we have computed genuine and imposter matches with Verifinger v12.0. In particular, we computed match scores (genuine and imposter) between the live impressions of the LivDet 2015 CrossMatch images as well as between the live samples of synthetic SpoofGAN fingerprints. These distributions are shown in Figure~\ref{fig:verifinger_hists} (a). This figure highlights that SpoofGAN is generating diverse fingerprint identities with similar intraclass and interclass variation as the real LivDet 2015 CrossMatch dataset, albeit producing slightly lower genuine scores compared to the real dataset. However, in terms of recognition performance at a fixed false acceptance rate (FAR), the performance between the two datasets is quite similar, despite the slightly shifted genuine distribution of the SpoofGAN images (see Table~\ref{table:tars}).

Furthermore, we computed genuine score distributions between the individual spoof types generated by SpoofGAN and their corresponding live impressions. These distributions are given in Figure~\ref{fig:verifinger_hists} (b). Here we see that Verifinger is able to successfully match spoof and live images of corresponding identities, which we believe opens the door to synthesising a large-scale spoof and recognition dataset that can be used to train and evaluate joint PAD and recognition algorithms.

\begin{table}
\centering
{\normalsize
\caption{TAR at Varying FAR Thresholds for LivDet 2015 CrossMatch Test Data vs. SpoofGAN Images.}
\begin{tabular}{ccc}
\noalign{\hrule height 1.5pt}
\textbf{FAR}            & \textbf{LivDet} & \textbf{SpoofGAN} \\
\noalign{\hrule height 1.0pt}
0.01\% @ threshold=48   & 99.87\%                        & 100\%                            \\
0.001\% @ threshold=60  & 99.80\%                        & 100\%                            \\
0.0001\% @ threshold=72 & 99.74\%                        & 99.80\%                          \\
1e-05\% @ threshold=84  & 99.41\%                        & 99.47\%                          \\
1e-06\% @ threshold=96  & 99.35\%                        & 98.87\%    \\
\noalign{\hrule height 1.5pt}
\end{tabular}
\label{table:tars}}
\end{table}

\subsubsection{Identity Leakage}
\begin{figure*}
\includegraphics[width=1.0\linewidth]{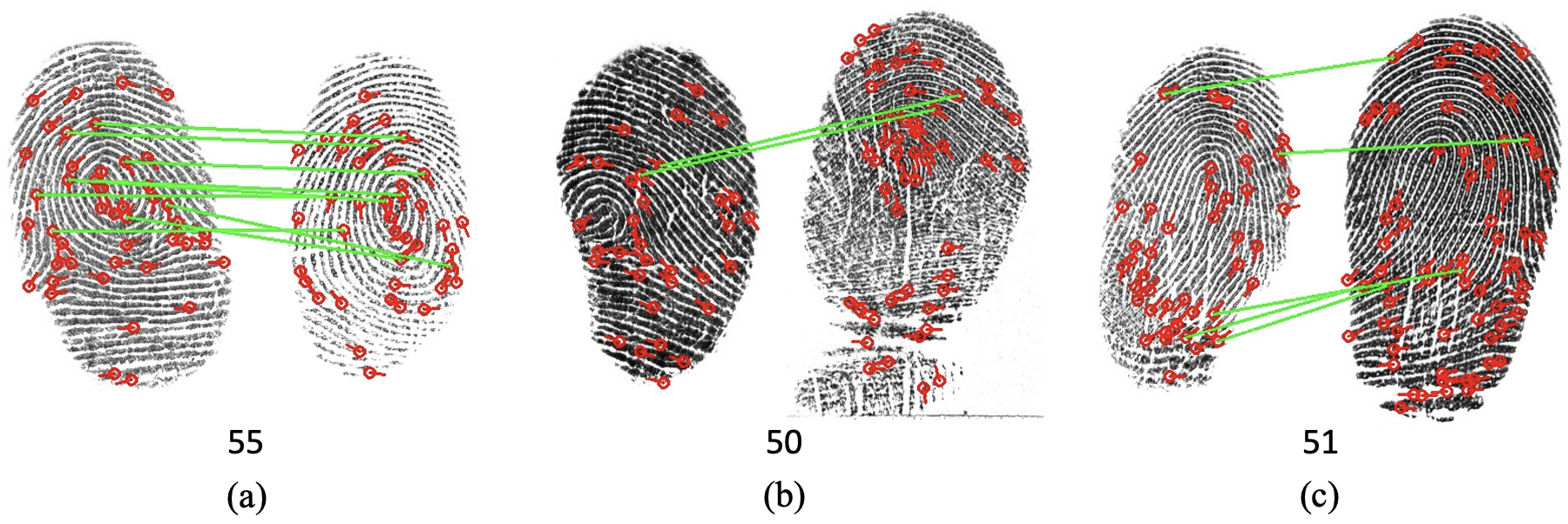} 
\caption{Example SpoofGAN and real training database image pairs with corresponding match scores given by Verifinger. For each pair, the left image is a SpoofGAN fingerprint image and the right image is a real fingerprint image. The threshold for a genuine match at an FAR of 0.01\% for Verifinger is 48, indicating that the identity of these training images have been ``leaked'' by SpoofGAN in the corresponding generated images.}
\label{fig:leakage}
\end{figure*}

A major advantage of generating synthetic fingerprint data is that, theoretically, no fingerprint identity belongs to an actual user somewhere in the world. However, there remains a concern that synthesis methods, such as GANs, may inadvertently over-fit and leak private information from the training corpus~\cite{bellovin2019privacy}. Therefore, it is instructive to investigate whether, and to what degree, any of our SpoofGAN generated images are revealing, i.e., match with sufficient confidence, the identities present in our training database. Toward this end, we have computed match scores between 1,500 SpoofGAN generated fingerprint identities and each of the 38,164 live fingerprint images in our real training set. Out of the roughly 57.2 million ($1,500\times38,164$) potential matches, only 50 comparisons exceeded the matching threshold of 48 set by Verifinger for a false acceptance rate of 0.01\% with a maximum match score of 81. Furthermore, the 50 matches resulted from just 29 SpoofGAN generated fingers out of the 1,500 evaluated. Some example matched SpoofGAN and real fingerprint image pairs are shown in Figure~\ref{fig:leakage} along with their corresponding match scores.

\subsection{Improved Spoof Classification with Synthetic Fingerprints}
Ultimately, our synthetic fingerprints should offer some utility in advancing the training of fingerprint spoof detection algorithms. Toward this end, we have augmented three existing, publicly available datasets of live and spoof fingerprints with our synthetic fingerprints in an effort to improve the performance beyond that achievable when training on the real live and spoof images from each dataset alone. For this evaluation, we have trained several spoof detection models on the following training set compositions: i.) synthetic live and spoof images only, ii.) real live and spoof images only, iii.) synthetic live and spoof images plus only real live images, and iv.) synthetic live and spoof images plus real live and spoof images. We used the SpoofBuster model, which consists of two Inception v3 networks, one trained on the whole image input and the other trained on $96\times96$ minutiae centered patches~\cite{chugh2018fingerprint}. The final spoof score is the weighted fusion of the two networks, with a minutiae patch score weight of 0.8 and a whole image score weight of 0.2. Each of the models are trained from scratch using Tensorflow on a single Nividia TitanX 1080 GPU on their respective datasets with identical hyper-parameters (learning rate of 0.01, step decay learning rate schedule, adam optimizer with default parameters, and total training updates of 200,000 steps). 

Shown in Table~\ref{tab:results}, we have evaluated each of the models on their respective test sets. Despite the lower performance when training on synthetic data alone compared to training on real data, we see improvement in the overall spoof classification performance when the real training data is augmented with samples from our synthetic live and spoof generator. For example, the error of the minutiae patch model trained on real data from LivDet 2013 is reduced by 91.03\% (from 15.60\% to 1.40\%) when augmented with synthetic data. Similarly, the error on LivDet 2015 is reduced from 0.48\% to 0.0\%, where the performance on GCT 6 remained the same at 100\%.

\begin{figure}
\includegraphics[width=1.0\linewidth]{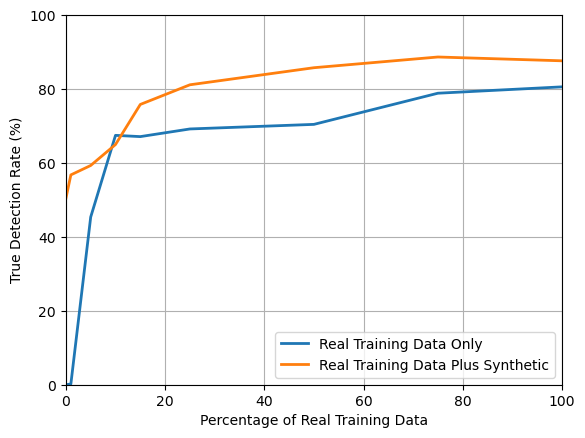} 
\caption{Spoof detection performance as we keep the number of synthetic training samples fixed and vary the percentage of real data included in training. Performance reported as TDR @ FDR = 0.2\% on the LivDet 2015 dataset.}
\label{fig:varying_percent}
\end{figure}

Lastly, it may be the case the researchers and practitioners have access to a large, private database of real live fingerprint images, whereas collecting an equivalent database of spoof fingerprints is more difficult and costly; therefore, they may wish to augment their database of real live images with synthetic spoof images. As seen in Table~\ref{tab:results}, mixing synthetic data with only real live fingerprint examples improves the performance over training on just the synthetic examples alone; however, synthetic spoof data is still not a substitute for collecting real spoof examples as the performance still lags quite significantly. Alternatively, Figure~\ref{fig:varying_percent} shows the trend in performance on LivDet 2015 as we keep the number of synthetic training samples fixed but varying the percentage of real data included when training the whole image-based spoof detection model. This figure suggests that when augmenting the training set with synthetic data, just 25\% of the original (real) data is required to obtain similar performance to training on 100\% of the real data alone, which significantly reduces the time and resources required for data collection to obtain similar performance.

\begin{table*}
{\normalsize
\caption{Spoof detection performance when trained on various combinations of real and synthetic data. We have followed the established protocols of train/test split for each of the evaluation datasets, which are shown in Table~\ref{tab:dataset_stats}}
\begin{tabularx}{\textwidth}{Z|Z|c|c||Z|Z|Z}
\noalign{\hrule height 1.5pt}
\multicolumn{4}{c||}{\textbf{Training Data Composition}}                                        & \multicolumn{3}{c}{\textbf{TDR @ 0.2\% FDR}}              \\
\hline
\textbf{Real Live} & \textbf{Real Spoof} & \textbf{Synthetic Live} & \textbf{Synthetic Spoof} & \textbf{LivDet2015} & \textbf{LivDet2013} & \textbf{GCT6} \\
\noalign{\hrule height 1.0pt}
                   &                     & \checkmark    & \checkmark                        & 36.53\%             & 42.30\%             & 56.08\%       \\
\hline
\checkmark    &                     & \checkmark          & \checkmark                    & 80.11\%             & 73.00\%             & 86.86\%       \\
\hline
\checkmark   & \checkmark             &                         &                          & 99.52\%             & 84.40\%             & \textbf{100.0\%}      \\
\hline
\checkmark     & \checkmark            & \checkmark     & \checkmark      & \textbf{100.0\%}            & \textbf{98.60\%}             & \textbf{100.0\%}     \\
\noalign{\hrule height 1.5pt}
\end{tabularx}
\label{tab:results}}
\end{table*}

\section{Conclusion and Future Work}
In this work, we presented a GAN-based synthesis method for generating high quality $512\times512$ plain fingerprint impressions of both live and spoof varieties. We demonstrated the utility of our synthetically generated spoof fingerprints in improving the performance of a spoof detector trained on real spoof fingerprint datasets. Additionally, our synthetic fingerprints closely resemble a database of real fingerprints both qualitatively and quantitatively in terms of various statistics, such as distribution of minutiae, NFIQ 2.0 image quality, spoof classification, and match score distributions computed by the state-of-the-art Verifinger 12.0 SDK. Finally, since our method is capable of generating multiple genuine and imposter fingerprints of unique identities of both live and spoof types, we open the door for large-scale training and evaluation of joint fingerprint spoof detection and recognition algorithms; overcoming a current limitation given the existing scale of publicly available spoof fingerprint datasets.

Despite the demonstrated utility of our synthetic spoof images, there remains several limitations that will be addressed in future work. First, given the lower standard deviation across the various fingerprint metrics presented in Table~\ref{tab:fp_stats}, the diversity of the generated images could be improved. A related issue, specific to fingerprint spoof detection, is the ever increasing novelty of spoof materials and types that may be encountered in the future; thus, instilling the generation process with the ability to adapt to novel spoof types is a promising future research direction. Similarly, leveraging synthetic data for improved cross-dataset and cross-material generalization is a promising future research direction. Lastly, training the synthetic fingerprint generator in an online fashion with a spoof detection network may provide additional supervision to generate more useful live and spoof examples to improve the spoof detection performance.

% \section*{Acknowledgment}

% Can use something like this to put references on a page
% by themselves when using endfloat and the captionsoff option.
\ifCLASSOPTIONcaptionsoff
  \newpage
\fi

\bibliography{cite}
\bibliographystyle{ieeetr}

\begin{IEEEbiography}[{\includegraphics[width=1in,height=1.25in,clip,keepaspectratio]{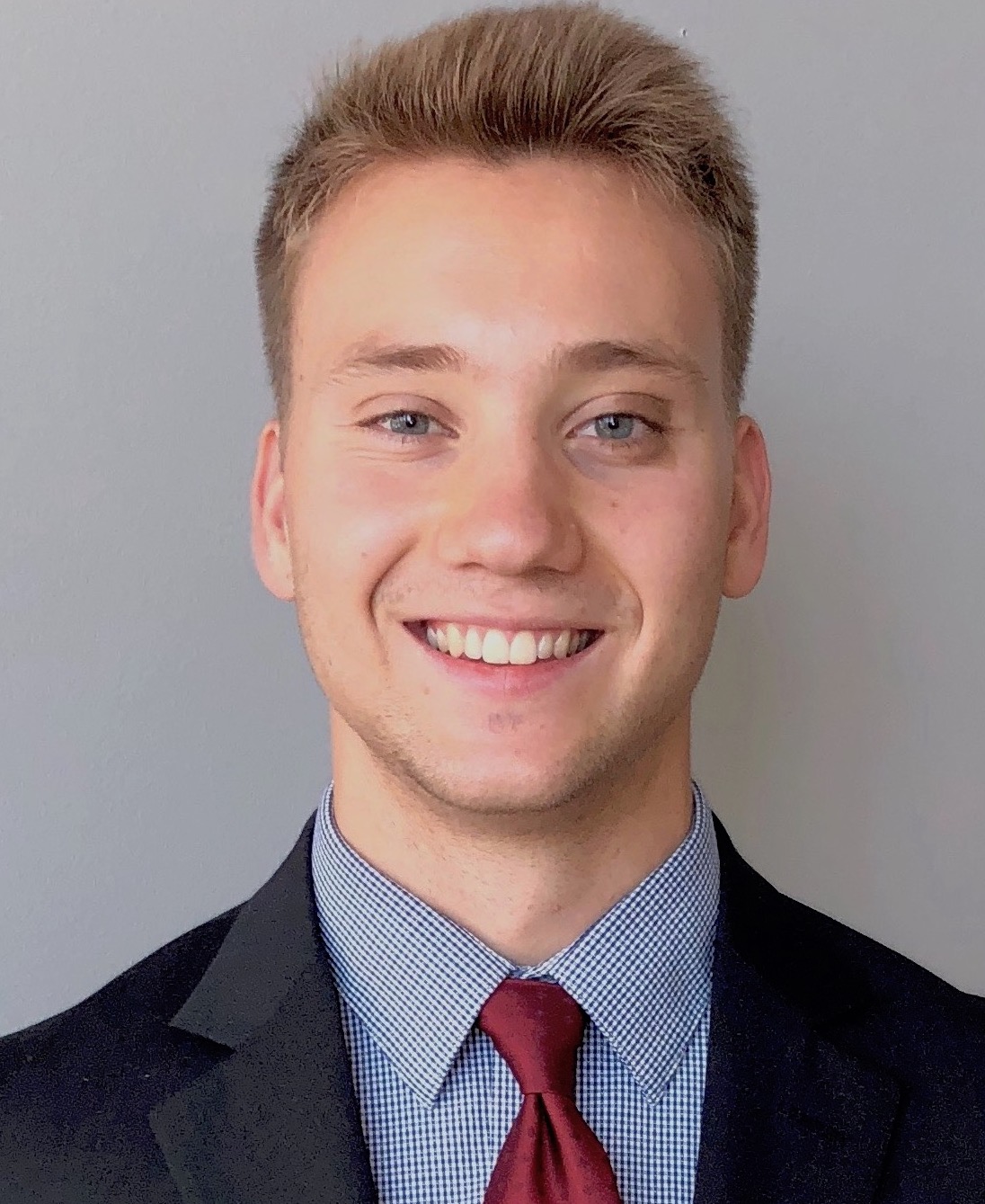}}]{Steven A. Grosz}
received his B.S. degree with highest honors in Electrical Engineering from Michigan State University, East Lansing, Michigan, in 2019. He is currently a doctoral student in the Department of Computer Science and Engineering at Michigan State University. His primary research interests are in the areas of machine learning and computer vision with applications in biometrics.
\end{IEEEbiography}

\begin{IEEEbiography}[{\includegraphics[width=1in,height=1.25in,clip,keepaspectratio]{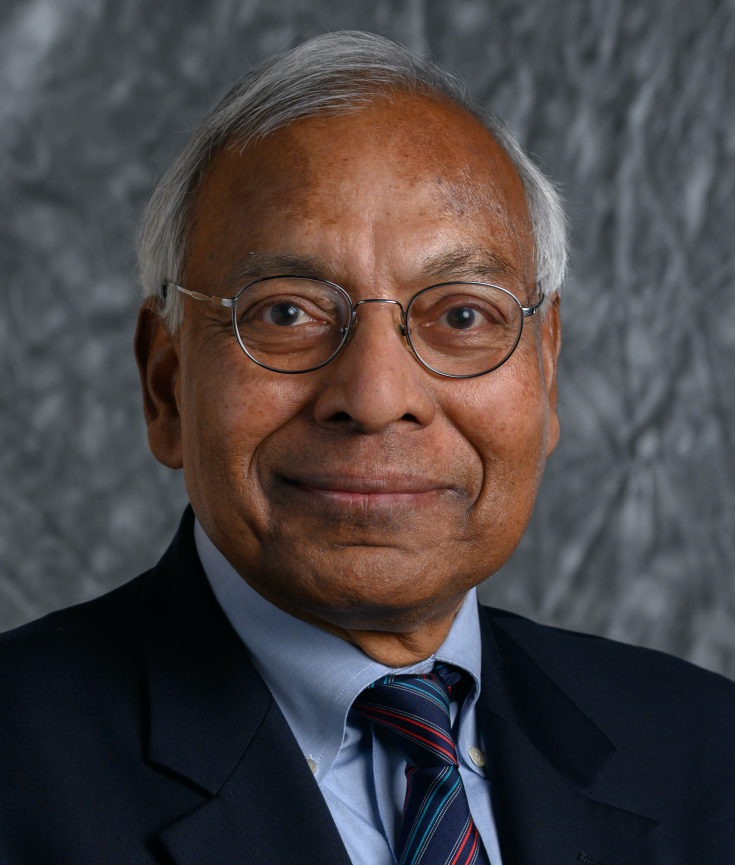}}]{Anil K. Jain}
Anil K. Jain is a University distinguished professor in the Department of Computer Science and Engineering at Michigan State University. His research interests include pattern recognition and biometric authentication. He served as the editor-in-chief of the IEEE Transactions on Pattern Analysis and Machine Intelligence and was a member of the United States Defense Science Board. He has received Fulbright, Guggenheim, Alexander von Humboldt, and IAPR King Sun Fu awards. He is a member of the National Academy of Engineering, the Indian National Academy of Engineering, the World Academy of Sciences, and the Chinese Academy of Sciences.
\end{IEEEbiography}

\end{document}